\documentclass[journal, twocolumn]{IEEEtran}
\IEEEoverridecommandlockouts

\usepackage{graphicx}
\usepackage{xcolor}
\usepackage{amsmath}
\usepackage{amssymb}
\usepackage{algorithm}
\usepackage{algorithmicx}
\usepackage[noend]{algpseudocode}
\usepackage{booktabs}
\usepackage{titlesec}
\usepackage{listings}
\usepackage{multirow}
\usepackage{placeins}
\usepackage{rotating}
\usepackage{pifont}
\usepackage{array}
\usepackage{mdframed}
\usepackage{lipsum}
\usepackage[utf8]{inputenc}
\usepackage[many]{tcolorbox}
\usepackage{varwidth}
\usepackage{colortbl}
\usepackage{array}
\usepackage[colorlinks=true, allcolors=blue]{hyperref}
\usepackage{cleveref}
\usepackage{makecell}
\usepackage{xcolor}

\usepackage{comment}
\usepackage{glossaries}
\usepackage{pgfplots}
\usepackage{adjustbox}
\usepackage{epstopdf}
\usepackage{subfigure}

\newacronym{UAV}{UAV}{Unmanned Aerial Vehicle}
\newacronym{VTOL}{VTOL}{Vertical takeoff and land}
\newacronym{EM}{EM}{Electromagnetic}
\newacronym{EMI}{EMI}{Electromagnetic Interference}
\newacronym{FW}{FW}{Fixed Wing}
\newacronym{SRTM}{SRTM}{Shuttle Radar Topography Mission}
\newacronym{TMI}{TMI}{Total Magnetic Intensity}

\begin{document}


\title{\Large Suspended Magnetometer Survey for Mineral Data Acquisition with Vertical Take-off and Landing Fixed-wing Aircraft}

\author{
Robel Efrem$^{\dagger}$, 
Alex Coutu$^{\star}$,  
Sajad Saeedi$^{\dagger}$
\vspace{-7 mm}
\thanks{\noindent$^{\dagger}$Toronto Metropolitan University\quad\quad $^{\star}$Rosor Exploration \newline \textcolor{white}{---} \{robel.efrem, s.saeedi\}@torontomu.ca, Alex@rosor.ca}%
}

\maketitle

\begin{abstract}
Multirotor Unmanned Aerial Vehicles (UAV)s have recently become an important instrument for collecting mineral data, enabling more effective and accurate geological investigations. This paper explores the difficulties in mounting high-sensitivity sensors on a UAV platform, including electromagnetic interference, payload dynamics, and maintaining stable sensor performance while in flight. It is highlighted how the specific solutions provided to deal with these problems have the potential to alter the collection of mineral data assisted by UAVs. The work also shows experimental findings that demonstrate the creative potential of these solutions in UAV-based mineral data collecting, leading to improvements in effective mineral exploration through careful design, testing, and assessment of these systems. 
\textcolor{black}{These innovations resulted in a platform that is quickly deployable in remote areas and able to operate more efficiently compared to traditional multirotor UAVs while still producing equal or higher quality mineral data. This allows for much higher efficiency and lower operating costs for high production UAV-based mineral data acquisition.}
\end{abstract}






\section{\textcolor{black}{Introduction}}\label{sec:introduction}

The urgent need for efficient mineral exploration is a driving force in the global economy, but traditional ground-based methods are hampered by labor intensity, time consumption, and environmental restrictions. This paper introduces an innovative solution to these limitations: the use of \gls*{UAV} based \gls*{VTOL} \gls*{FW} systems for mineral data acquisition. Leveraging \gls*{UAV}s enables rapid coverage of large and remote areas, a crucial advantage in the fast-paced and ever-evolving field of mineral exploration. This approach not only accelerates the process but also reaches terrains that are otherwise inaccessible, marking a significant advancement in the field.

The integration of high-sensitivity sensors on \gls*{UAV} platforms for mineral exploration presents unique challenges \cite{mohsan2023editorial}. These include managing payload limitations, dynamics, and safety, particularly in the case of fixed-wing \gls*{UAV}s \cite{vangu2022use}. The most significant obstacle is minimizing \gls*{EMI} while ensuring the suspended magnetometer systems maintain high data quality during flight \cite{shahsavani2021aeromagnetic}. These challenges stem from the complex interplay of UAV design constraints, sensor sensitivity, and the need for operational stability in varied and often harsh environmental conditions~\cite{leech2021acquisition}.

Recent trends in \gls*{UAV}-based mineral exploration have seen a focus on \gls*{VTOL} fixed-wing \gls*{UAV}s, lauded for their increased flight endurance, flexibility, and precision \cite{tziavou2018unmanned}. However, most studies and applications have not delved deeply into mineral sensing applications with a suspended payload or addressed the specific challenges of \gls*{EMI} in detail \cite{lu2023development}. This gap in research and application highlights a significant opportunity for innovation. The current state-of-the-art solutions, while effective for broader applications like topographic mapping, fall short in addressing the nuanced requirements of mineral sensing, particularly in terms of \gls*{EMI} management and payload dynamics optimization \cite{giordan2020use}.

\begin{figure}[t!]
\vspace{2 mm}
    \centering
    \includegraphics[width=\linewidth]{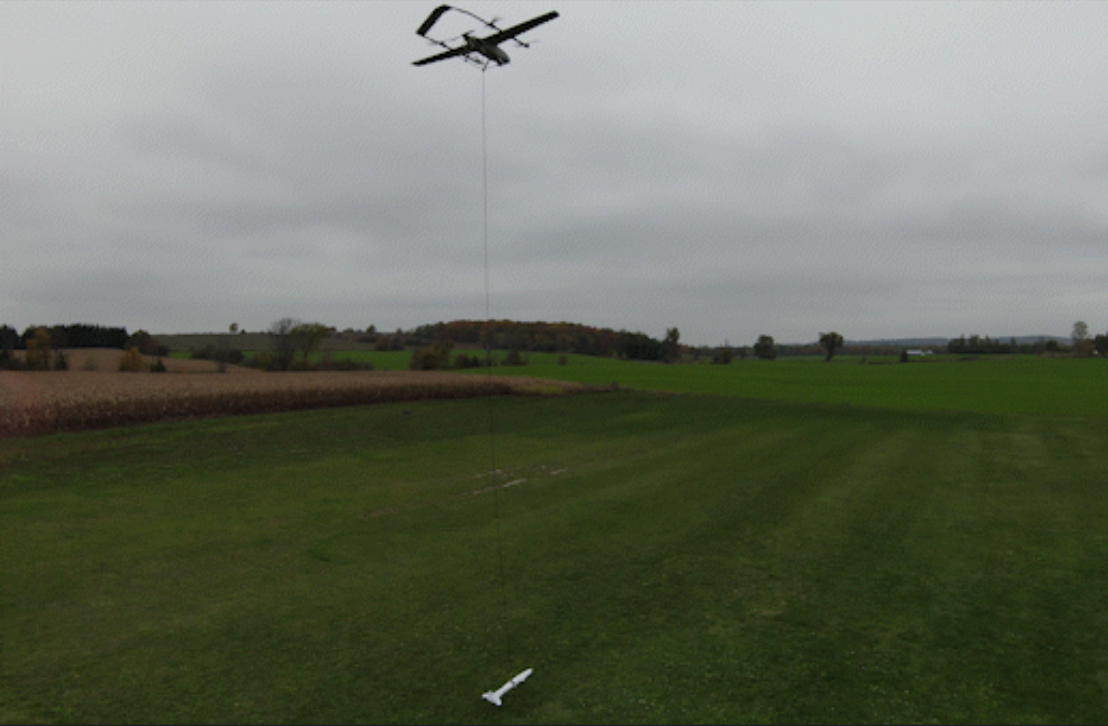}
    \caption{Vertical take-off and landing (VTOL) fixed-wing aircraft taking off with the suspended payload.}
    \label{fig:sys}
\end{figure}

This research aims to bridge these gaps by enhancing \gls*{VTOL} \gls*{UAV} capabilities for mineral exploration. We focus on reducing electromagnetic interference in suspended magnetometer setups and maintaining an acceptable level of data quality. Our approach involves extensive testing and refinement of \gls*{UAV} design, including structural modifications for payload stability and innovative tethering methods to minimize \gls*{EMI}. Experimental findings demonstrate the effectiveness of these solutions in producing high-quality, reliable mineral data. Our contributions lie in the development of a UAV platform that is not only quickly deployable and efficient but also capable of operating in challenging terrains while ensuring superior data quality and reduced operational costs. This integrated system marks a notable advancement in UAV-based mineral data acquisition, setting new standards in the field (See Fig~\ref{fig:sys}). To better illustrate how this system differs compared to existing surveying systems, Table~\ref{tab:comparison} outlines a general comparison between UAVs and manned surveying techniques. Our proposed method has relatively low cost  while being able to provide better or similar results compared to the crewed options. See videos of the proposed system on the website of the project\footnote{\href{https://sites.google.com/view/mineral-aircraft/vtol-fixed-wing-uav}{https://sites.google.com/view/mineral-aircraft/vtol-fixed-wing-uav}}.

\begin{table*}[h!]
    \centering
    \definecolor{Gray}{gray}{0.9}
    \begin{adjustbox}{max width=\textwidth}
    \begin{tabular}{|l|c|c|c|c|}
        \toprule
        \textbf{Metric} & \textbf{Generic \gls*{UAV}} & \textbf{\gls*{VTOL} FW \gls*{UAV} (Ours)} & \textbf{Fixed Wing (Crewed)} & \textbf{Helicopter (Crewed)} \\
        \midrule
        Cost of Aircraft (\$) & 20,000 & \cellcolor{Gray}17,000 & 1,500,000 & 2,100,000 \\
        Flight Duration (hrs) & 0.5 & 3 & \cellcolor{Gray}6 & 3 \\
        Safety (for humans, scale 1-10) & \cellcolor{Gray}10 & \cellcolor{Gray}10 & 8 & 7 \\
        Coverage Area per Flight (km\(^2\)) & 0.5 & 5.9 & \cellcolor{Gray}60 & 15 \\
        Typical line spacing (m) & \cellcolor{Gray}50 & \cellcolor{Gray}50 & 100 & 100 \\
        Cost per Line Km (\$/km) & 1.38 & \cellcolor{Gray}0.78 & 5.00 & 10.00 \\
        Flight Time per km (seconds/km) & 90 & 45 & \cellcolor{Gray}18 & 35 \\
        Average Operating Costs (\$ per day) & \cellcolor{Gray}500 & \cellcolor{Gray}500 & 8,000 & 12,000 \\
        \bottomrule
    \end{tabular}
    \end{adjustbox}
    \caption[General comparison between mineral sensing platforms]{\textcolor{black}{General comparison between various \gls*{UAV}s, manned fixed wing, and manned helicopter for mineral sensing. Highlighted areas indicating the favorable value for each category. Platform comparison specifications are taken from that of a DJI m300, 
    our custom \gls*{VTOL} \gls*{FW} \gls*{UAV}, a Piper PA-31 Navajo, and a Bell 206 JetRanger respectively. Metric calculations can be found in Appendix~\ref{appendixA}}.}
    \label{tab:comparison}
\end{table*}
\section{\textcolor{black}{Literature Review}} \label{sec:literature}
This literature review synthesizes key findings from existing works on \gls*{UAV}s in the context of mineral sensing, particularly focusing on \gls*{VTOL} fixed-wing \gls*{UAV}s and their applications in mineral exploration.

Ground magnetic surveying, a long-standing technique in geophysics, involves walking with a magnetometer to measure surface magnetic fields, useful for detecting shallow subsurface features like metallic objects or mineral deposits \cite{vitale2019new}. Though less expensive and flexible in survey area, it's time-consuming and challenging, especially over large, difficult terrains, often requiring high-resolution follow-up scans to confirm mineral deposits. Aerial surveying, utilizing sensor-equipped aircraft, overcomes these limitations by quickly capturing mineral data over large areas. Commonly, this involves attaching a magnetometer to fixed-wing aircraft, preferred for their ability to cover vast distances quickly at high altitudes, albeit at the cost of reduced data resolution and maneuverability in complex terrains \cite{mat2023magnetic}. Helicopters offer an alternative with their lower altitude flight and better maneuverability, yielding higher resolution data ideal for smaller or more rugged areas, though they are generally more costly and less efficient for large-scale surveys \cite{persova2021resolution}.

The emergence of \gls*{UAV} for aerial surveying represents a significant advancement, especially in terms of data resolution. Drones, being smaller and capable of flying at lower altitudes, are well-suited for small area surveys or those involving complex terrains \cite{jiang2020integration}. While offering higher resolution data and improved safety by minimizing the need for manned flights, drones face limitations in flight time, control range, and payload capacity \cite{greengard2019when}. Despite these constraints, the development of \gls*{UAV}s capable of high endurance and long-distance flights at low altitudes offers a promising direction. These systems could provide a more efficient, cost-effective means of acquiring high-resolution mineral data over larger areas, addressing both the limitations of traditional ground surveys and the challenges faced by manned aerial survey platforms.

Advancements in \gls*{UAV} technology have significantly impacted the mineral sensing domain. The use of \gls*{VTOL} fixed-wing \gls*{UAV}s has gained attention for their potential in mineral exploration, offering notable advantages such as improved flight endurance, deployment flexibility, and precise positioning capabilities in varied terrains \cite{nex2014uav, akshat2022review}. These features are critical in tasks requiring access to remote locations and high-resolution data acquisition \cite{padua2017uas}. Despite these benefits, a gap in research is evident, as most studies lack detailed exploration of \gls*{VTOL} fixed-wing \gls*{UAV}s' specific use in mineral sensing applications.

The integration of advanced sensor technologies in \gls*{UAV}s has been pivotal in enhancing mineral exploration methods. Hyperspectral imaging and LiDAR have emerged as popular choices for their detailed surface analysis and topographic mapping capabilities, enabling the differentiation of rock types and identification of geological features \cite{barton2021extending, niethammer2012uav, salvini2015geological}. The combination of multiple sensors, such as hyperspectral with magnetic sensors or LiDAR with multispectral imaging, has further expanded the \gls*{UAV}s' capabilities in mineral detection \cite{jackisch2019drone, shendryk2020fine}.

Fixed-wing \gls*{UAV}s are often preferred in mineral sensing for their extended flight times and higher payload capacities, whereas multirotor \gls*{UAV}s offer better maneuverability at lower airspeeds \cite{dundar2020design, shahmoradi2020comprehensive}. However, the choice between different \gls*{UAV} platforms depends on specific mission requirements, such as survey scale and data resolution needs.

While \gls*{UAV}-based mineral sensing presents numerous advantages, challenges remain in signal-to-noise ratio, resolution, calibration, and data fusion, especially when employing hyperspectral and LiDAR sensors \cite{goetz2009three, vosselman2013recognising}. Additionally, integrating multiple sensors on a single \gls*{UAV} platform poses difficulties in data interpretation and fusion \cite{thiele2021multi}. The regulatory landscape also plays a significant role in \gls*{UAV} deployment for mineral exploration, with varying legal frameworks across jurisdictions \cite{molnar2016unmanned}.

In conclusion, the current literature underscores the potential of \gls*{VTOL} fixed-wing \gls*{UAV}s in mineral sensing, highlighting the need for further research into their application-specific optimization. Future studies should also focus on comparing operational costs, maintenance requirements, and environmental impacts of these \gls*{UAV} systems. Understanding these aspects will be crucial in developing more specialized and efficient \gls*{UAV}-based solutions for mineral exploration.

\begin{figure*}[t!]
    \centering
    \includegraphics[width=12.5cm]{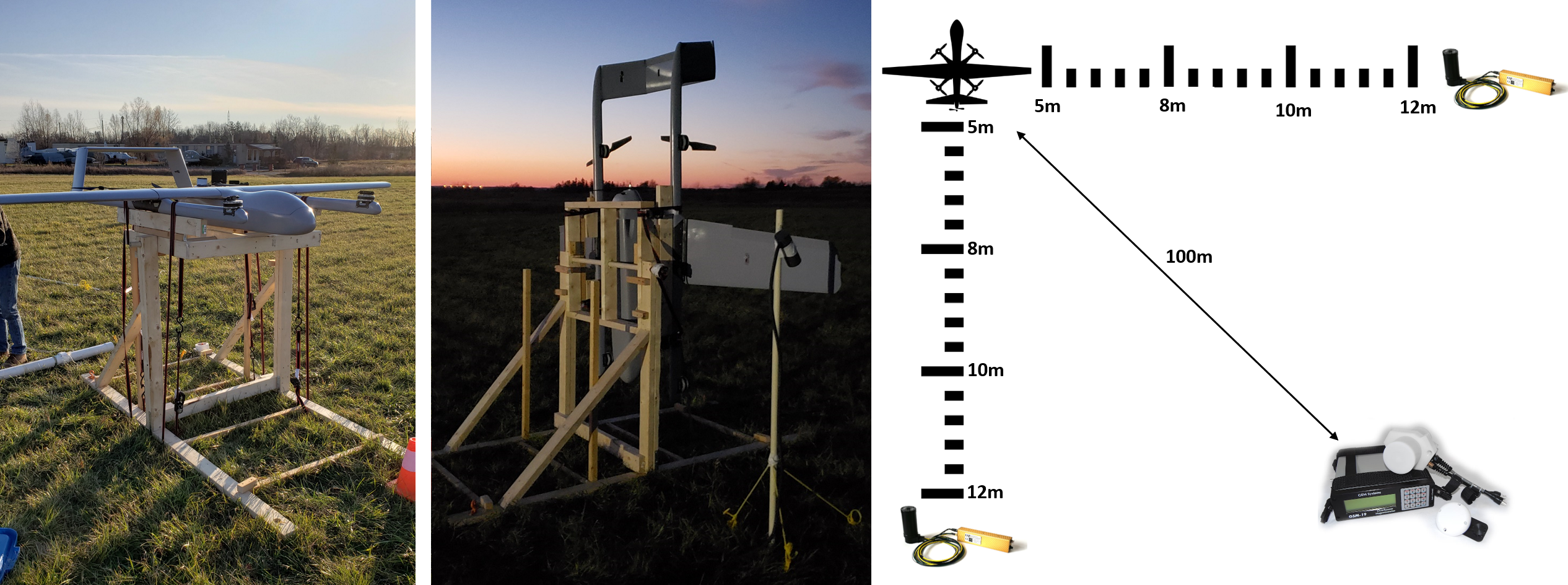}
    \caption{(Left) VTOL aircraft mounted to EMI measurement frame. The magnetometer sensor is on a stand to the right of the aircraft \textcolor{black}{(shown in the picture on the middle).} {\textcolor{black}{(Right) Diagram of how the sensors were placed with respect to the drone for EMI testing. The GSMP-35U Potassium Magnetometer (top right and bottom left) was placed between 5-12m from the drone and the GSM-19 Overhauser magnetometer (bottom right) was placed 100m away as a neutral reference}} }
    \label{fig:Magvtol}
\end{figure*}

\section{Proposed Method}\label{sec:method}
\textcolor{black}{
In the upcoming section, we will delve into proposed strategies aimed at tackling issues such as electromagnetic interference and payload dynamics. These strategies will underscore the capacity to revolutionize mineral data acquisition with the assistance of Unmanned Aerial Vehicles (UAVs) and yield substantial advancements in the wider realm of mineral exploration.
}


\subsection{Sensor Integration}
To effectively integrate a suspended magnetometer payload to a \gls*{VTOL} fixed-wing aircraft, first, an examination of the optimal tether length must be conducted which would minimize EMI interference while maintaining safe operation. \textcolor{black}{In this case the main source of the \gls*{EMI} is produced by the high current forward propulsion motor and its accompanying electronic speed controller onboard the aircraft.} The tether plays a crucial role in minimizing EMI interference between the \gls*{UAV} and the magnetometer sensor. A series of experiments were conducted to determine the optimal tether length for the suspension system. These tests involved strapping the full system with the propellers inverted to a frame secured to the ground, then running the electric propulsion system at different levels with a ground-based magnetic sensor recording the noise levels. All of the testing was performed at dusk when the diurnal variations caused by the Sun's magnetic field and solar wind are at their lowest \cite{egbert2020modelling}. A photograph of the setup is shown in 
Fig.~\ref{fig:Magvtol} (left and middle). 
A ground magnetometer was also placed 100m away from the testing site as well as any roadways or power lines to provide a reference for the testing data should there be any geomagnetic storms {causing} any false positives in the data. The main magnetic sensor was placed in different areas with respect to the drone as well as different distances. The tests were repeated once again with the drone oriented vertically and spinning the motors at different rpms to ensure a wide variety of points in 3d space relative to the drone were recorded. \textcolor{black}{A diagram of the testing procedure is shown in 
Fig.~\ref{fig:Magvtol}-(right.}


\textcolor{black}{After processing the results the noise emitted from the main rear-mounted propulsion motor for the fixed wing configuration only outputted detectable noise up to about 9 meters at full throttle. Due to the EMI noise spectrum following the inverse square law
the intensity of the EMI signal is inversely proportional to the square of the distance from the source and the intensities recorded corroborate this notion \cite{inverse-square-law-2014, Halliday2021}. \textcolor{black}{The motor used for testing was a T-motor AT7224 40CC which can output approximately 17kg of thrust while drawing 6kw of power at peak thrust, to sustain flight however only about 1.5kw is typically required \cite{tmotor2023at7224}.} Regardless of the power level, at 10 meters distance from the \gls*{UAV} there is virtually no discernible difference to the magnetic readings while the \gls*{UAV} is at full power or shut down.}


\subsection{UAV and Magnetometer Modifications}

\textcolor{black}{Adapting the fixed-wing \gls*{VTOL} \gls*{UAV} platform with a suspended magnetometer payload required a number of structural and functional modifications, \textcolor{black}{ these changes included the following list of hardware modifications:
\begin{enumerate}
    \item \textbf{Airframe Reinforcement:} Strengthening of the airframe to withstand forces exerted by the suspended payload during various flight conditions.
    \item \textbf{Payload Mounting Point:} Integration of a large carbon fiber plate with U bolt at the center of gravity of the aircraft.
    \item \textbf{Safeguard Mechanism for Propeller:} Implementation of a cage like mechanism to prevent tether entanglement in the UAV's forward flight propeller during transition phases and flight.
    \item \textbf{Aerodynamic and Propulsion Adjustments:} Modifications to the UAV’s aerodynamics and propulsion system to support the suspended payload, maintaining flight stability and efficiency.
    \item \textbf{Specialized Payload Mounting Brackets:} Design of non-ferrous mounting brackets to securely attach the payload to the tether.
    \item \textbf{Aerodynamic Payload Cover:} Development of an aerodynamic cover for the payload, featuring an empennage for self-stabilization during high-speed flights.
\end{enumerate}
}
To begin with, the airframe needed to be reinforced to withstand the forces exerted by the suspended payload during various flight conditions. The mounting point of the suspended payload was connected to a large carbon fiber plane intended to spread the load along the fuselage while still centering the force at the center of gravity. It is crucial that a slug payload is mounted at the center of gravity of a fixed-wing platform to ensure disturbances in the payload do not cause attitude alterations while in flight. Additionally, a safeguard mechanism was implemented that prevented the tether from getting entangled in the \gls*{UAV}'s propellers during the transition,  while in flight, and de-transition. Additional adjustments were made to the \gls*{UAV}'s aerodynamics and propulsion system to better accommodate the suspended payload without compromising flight stability and efficiency.}

The magnetometer payload required modifications to ensure compatibility with the suspension system and maintain aerodynamic stability during flight. Specialized mounting brackets were designed to securely attach the payload to the tether at a single point without the use of any ferrous materials that may alter the magnetic readings. Furthermore, an aerodynamic cover was developed for the payload, which not only protected the sensor from environmental factors but also utilized an empennage to help self-stabilize the payload during high-speed flights. This cover reduced drag and turbulence effects on the payload, allowing for more accurate and reliable data acquisition without allowing the suspended payload to tumble in the wind.

\subsection{Cable Length for Suspended Payload}

In order to properly estimate the required cable length for the suspended payload from a dynamics perspective, an equation needs to be formulated taking into account all of the necessary attributes of the system. The intent is to design a system that when flown produces a suspended payload system with the least possible amount of swing as well as settling time. The equation is a simplified model and does not account for several factors such as aerodynamic effects, wind conditions, and complex interactions between the drone and payload. A more precise estimation of the system can be achieved using processes such as computational fluid dynamics simulations to validate and refine the model but instead, practical experiments will be completed based on the mathematical output.

\begin{figure}[t!]
    \centering
    \includegraphics[width=8cm]{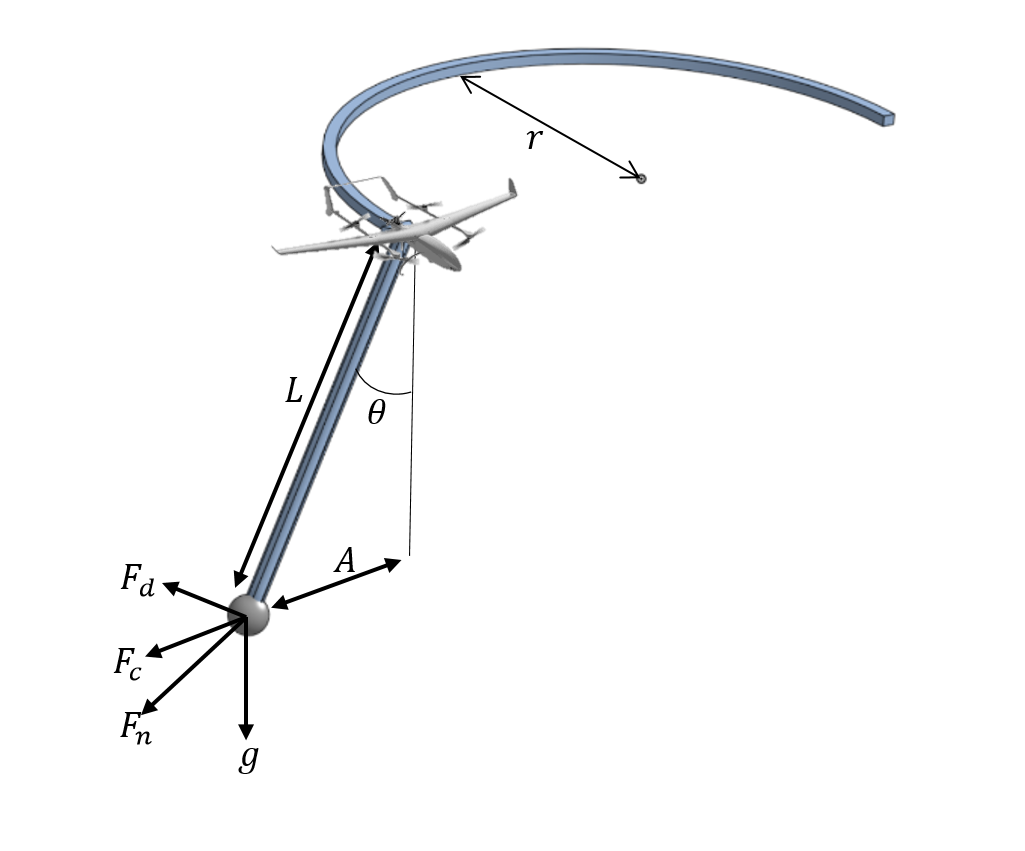}
    \caption[Free body diagram representing the payload dynamics]{\textcolor{black}{Free body diagram representing the payload dynamics of the system, the point at the base representing the payload and the curve representing the path of the UAV}}
    \label{fig:FBD}
\end{figure}

To develop an appropriate {relation} to represent this system, first, the algorithm variables and descriptions must be outlined (see Fig.~\ref{fig:FBD}):
\begin{itemize}
  \item $A$: the amplitude of the swing (meters).
  \item $m_p$: the mass of the payload (kg).
  \item $V$: the speed of the drone (m/s).
  \item $r$: the radius of the turn (m).
  \item $g$: the gravitational acceleration ($m/s^2$, typically 9.81 $m/s^2$).
  \item $L$: the length of the rope (m).
  \item $\zeta$: the damping ratio of the rope material (dimensionless).
  \item $C_d$: the drag coefficient of the payload (dimensionless).
  \item $A_p$: the cross-sectional area of the payload (m²).
  \item $\rho$: the density of air (kg/m³).
\end{itemize}

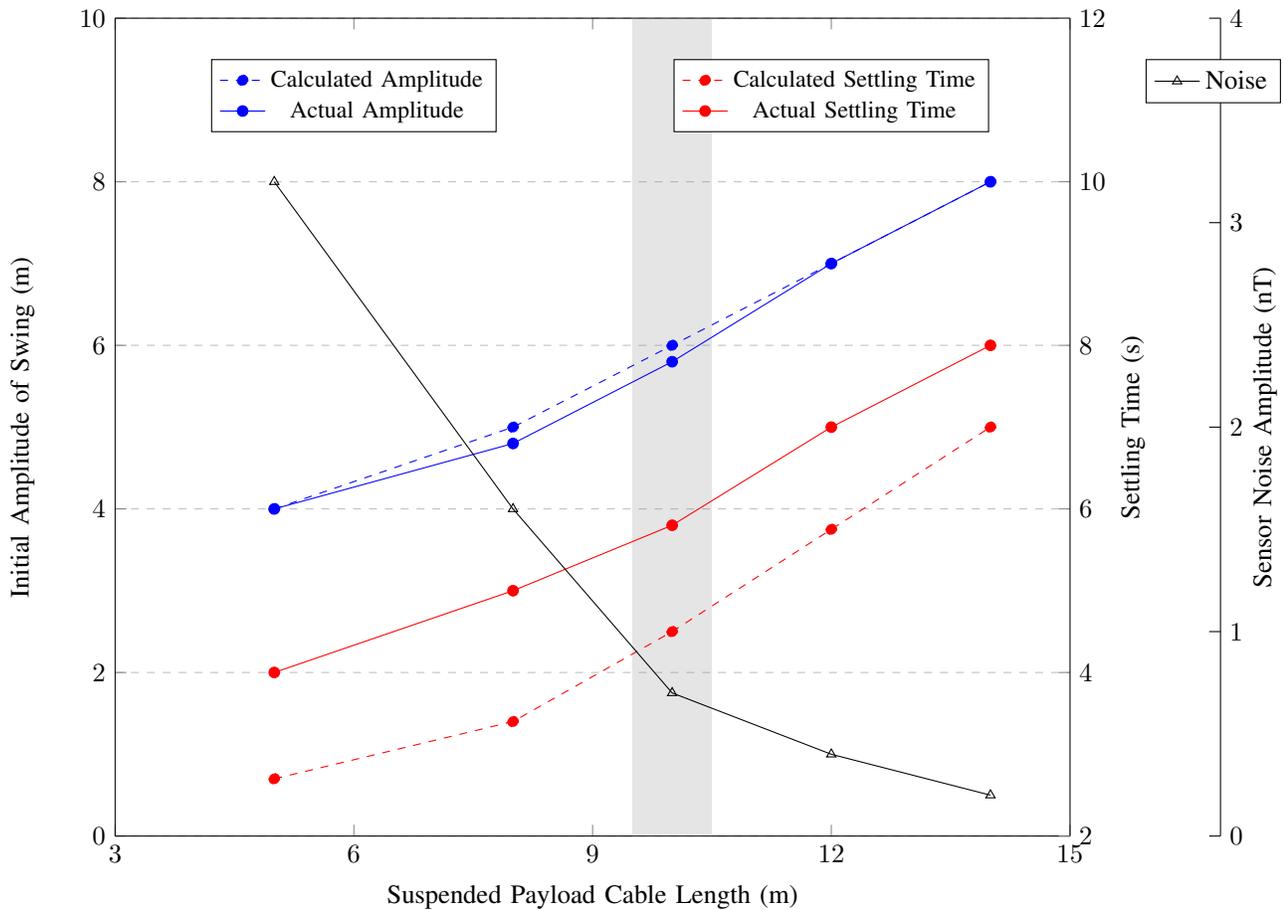
\begin{figure*}[t!]

\begin{tikzpicture}
\pgfplotsset{every axis/.style={ymin=0}}
\begin{axis}[
    scale only axis,
    xmin=3, xmax=15,
    xlabel={Suspended Payload Cable Length (m)},
    ylabel={Initial Amplitude of Swing (m)},
    ymin=0, ymax=10,
    xtick={3,6,9,12,15},
    ytick={0,2,4,6,8,10},
    ymajorgrids=true,
    grid style=dashed,
    axis y line*=left,
    width=0.7\textwidth, height=0.6\textwidth,
    legend style={font=\small, at={(0.25,0.95)}, anchor=north}
]

\fill [gray, opacity=0.2] (axis cs: 9.5, 0) rectangle (axis cs: 10.5, 10);

\addplot[
    color=blue,
    mark=*,
    dashed,
]
coordinates {
    (5,4)(8,5)(10,6)(12,7)(14,8)
};
\addlegendentry{Calculated Amplitude}

\addplot[
    color=blue,
    mark=*,
]
coordinates {
    (5,4)(8,4.8)(10,5.8)(12,7)(14,8)
};
\addlegendentry{Actual Amplitude}

\end{axis}

\begin{axis}[
    scale only axis,
    xmin=3, xmax=15,
    axis y line*=right,
    axis x line=none,
    ylabel={Settling Time (s)},
    ylabel style={yshift=-42em},
    ymin=2, ymax=12,
    ytick={2,4,6,8,10,12},
    xtick=\empty,
    width=0.7\textwidth, height=0.6\textwidth,
    legend style={font=\small, at={(0.75,0.95)}, anchor=north}
]

\addplot[
    color=red,
    mark=*,
    dashed,
]
coordinates {
    (5,2.7)(8,3.4)(10,4.5)(12,5.75)(14,7)
};
\addlegendentry{Calculated Settling Time}

\addplot[
    color=red,
    mark=*, 
]
coordinates {
    (5,4)(8,5)(10,5.8)(12,7)(14,8)
};
\addlegendentry{Actual Settling Time}

\end{axis}

\begin{axis}[
    scale only axis,
    xmin=3, xmax=15,
    axis y line*=right,
    axis x line=none,
    ylabel={Sensor Noise Amplitude (nT)},
    ylabel style={yshift=-47em},
    ymin=0, ymax=4,
    ytick={0,1,2,3,4},
    xtick=\empty,
    width=0.7\textwidth, height=0.6\textwidth,
    every outer y axis line/.style={xshift=2cm},
    every tick/.style={xshift=2cm},
    every y tick label/.style={xshift=2cm},
    legend style={at={(1.15,0.95)}, anchor=north}
]

\addplot[
    color=black,
    mark=triangle,
]
coordinates {
    (5,3.2)(8,1.6)(10,0.7)(12,0.4)(14,0.2)
};
\addlegendentry{Noise}

\end{axis}
\end{tikzpicture}
\caption[Payload dynamics comparison]{\textcolor{black}{Payload Dynamics Comparison: Calculated vs. Actual vs. Sensor Noise. The Grey zone indicates the configuration with the best compromise between payload dynamics and sensor noise.}}
\label{fig:length}
\end{figure*}

\noindent The algorithm can be formulated by first calculating the centrifugal force ($F_c$) on the payload during the turn using \textcolor{black}{Eq.~\eqref{eq:1}}. Then calculate the change in the angle ($\theta$) of the rope due to the centrifugal force (Eq.~\eqref{eq:2}) as well as aerodynamic drag on the payload ($F_d$) (Eq.~\eqref{eq:3}) and finally the net force ($F_n$) on the payload due to product of the three forces (Eq.~\eqref{eq:4})~\cite{Halliday2021}:
\begin{equation}
F_c = \frac{{m_p V^2}}{{r}}
\label{eq:1}
\end{equation}
\begin{equation}
\theta = \arctan \left( \frac{{F_c}}{{m_p  g}} \right)
\label{eq:2}
\end{equation}
\begin{equation}
F_d = \frac{1}{2} C_d  A_p \rho V^2
\label{eq:3}
\end{equation}
\begin{equation}
F_n = \sqrt{F_c^2 + (m_p  g)^2 + F_d^2}
\label{eq:4}
\end{equation}
The natural frequency ($\omega_n$) of the payload swing (Eq.~\eqref{eq:5}) as well as the damping coefficient ($c$) of the rope (Eq.~\eqref{eq:6}) can be calculated based on the system specifications~\cite{Halliday2021}:
\begin{equation}
\omega_n = \sqrt{{\frac{{g}}{{L}}}}
\label{eq:5}
\end{equation}
\begin{equation}
c = 2 m_p \zeta  \omega_n
\label{eq:6}
\end{equation}
The amplitude of the swing ($A$) can then be estimated using an approximation for a damped harmonic oscillator. After substituting in all the necessary variables, we have the final equation for amplitude of swing from a suspended payload system of a fixed-wing \gls*{UAV} shown in Eq.~\eqref{eq:7}: 
\begin{equation}
A \approx \frac{{F_n / m_p}}{{\sqrt{{(\omega_n^2 - (V^2 / r^2))^2 + \left(\frac{{c V}}{{m_p r}}\right)^2}}}}
\label{eq:7}
\end{equation}
To calculate the settling time, the effective damping coefficient ($c_{\text{eff}}$) must first be established, considering both the inherent damping and the damping due to drag shown in Eq.~\eqref{eq:8}. The settling time ($T_s$) for the payload can be approximated by using the settling time equation for an underdamped harmonic oscillator. With ($A$) in this case being the initial amplitude from the above equation and ($A_{final}$) being the final amplitude within which the oscillation must remain (Eq.~\eqref{eq:9}).
\begin{equation}
c_{\text{eff}} = c + \frac{1}{2} C_d  A_p  \rho  V
\label{eq:8}
\end{equation}
\begin{equation}
T_s \approx \frac{\ln\left(\frac{A}{A_{\text{final}}}\right)}{\frac{{c_{\text{eff}}}}{{2  m_p}}  \omega_n}
\label{eq:9}
\end{equation}
After inputting all the necessary system characteristics, the payload swing amplitude and settling time can be plotted to determine if the expected system can perform favorably at the EMI limit of 10m. The best compromise between payload swing amplitude/settling time and EMI will be selected as the final system suspension length after testing (See Fig.~\ref{fig:length}).

\noindent With a 10-meter cable length selected as the target, the next stage of testing involves performing a \textcolor{black}{full systems test} magnetic survey with a suspended magnetometer to prove the data is usable. Various other lengths including 8m and 12m were tested in controlled environments using a fixed-wing \gls*{VTOL} \gls*{UAV} platform. \textcolor{black}{This testing not only verifies that the suspended payload system will work to produce high quality magnetic data set, but also which cable length provides the best mix between low EMI noise and acceptable payload dynamics.} As the payload swings beneath the \gls*{UAV} in forward flight, this introduces a new mechanical noise into the system where the payload is taking measurements a few meters off from where it should be producing a lower-quality dataset. As the cable length increases the frequency of the payload swing decreases while the period and amplitude increase. Normally in a pendulum system, the amplitude would decrease with an increase in length but in this case, the swing is caused primarily by the initial turn at the start and end of each flight line.

\begin{figure*}[t!]
    \centering
    \includegraphics[width=\textwidth]{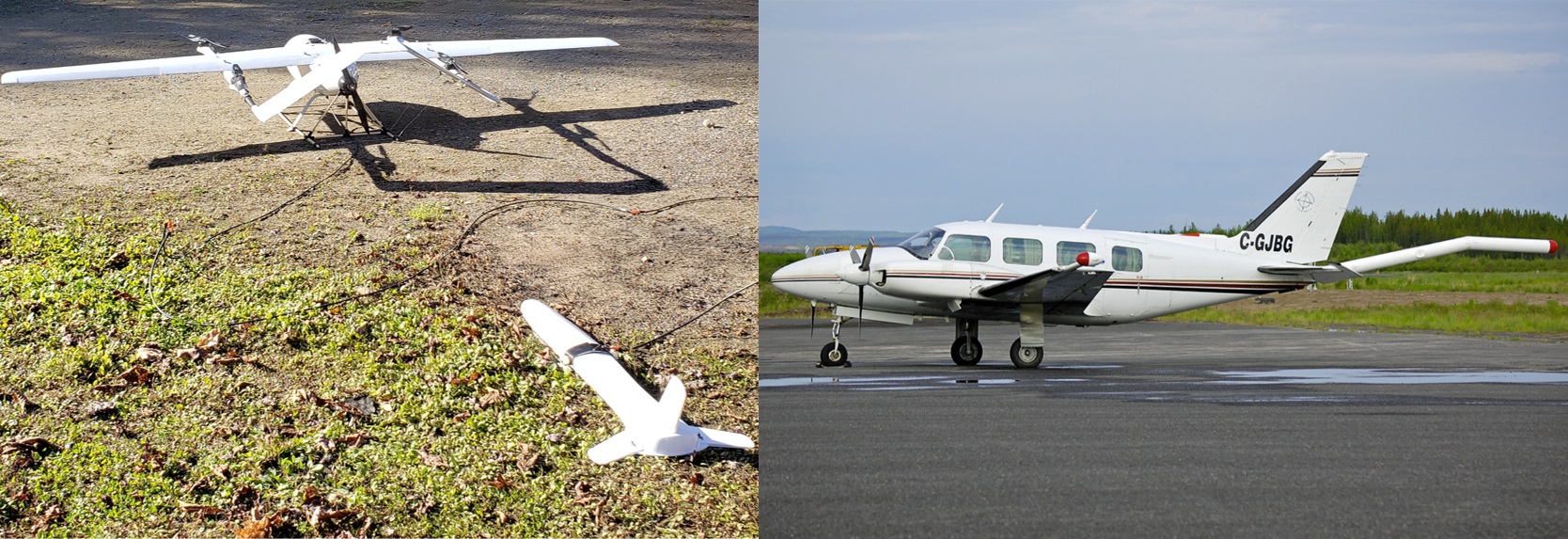}
    \caption[Fixed-wing \gls*{VTOL} UAV Vs fixed-wing manned]{\textcolor{black}{Fixed-wing \gls*{VTOL} UAV with towed bird style magnetometer, in this case, a Geometrics MagArrow (Left) and fixed-wing manned aircraft (Piper Navajo) with rear mounted stinger magnetometer (right). ~\cite{willems2009}}}
    \label{fig:PiperVSdrone.jpg}
\end{figure*}

Measuring the amplitude and settling time of the suspended payload under the \gls*{UAV} was achieved through the use of the drone's onboard camera. The camera, aimed at the payload, captured visual data of the payload's movements, and post-processing of this data allowed for the evaluation of payload dynamics. The amplitude, or maximum swing of the payload, was determined by identifying the furthest points of the payload's swing in the captured footage and measuring the pixel distance between them. This pixel distance was converted into the actual distance by using the known measurements of the payload size. The settling time was calculated by identifying the point in the video where the payload starts swinging (caused by initial turns at the start and end of each flight line) and the point where the swing amplitude falls within an acceptable range. The time difference between these two points is the settling time.

With a long cable length the payload will swing much higher causing the initial amplitude to be greater, this coupled with a longer settling time produces unfavorable results for magnetic sensing. The acquired data was analyzed for all types of noise levels, considering the impact of EMI interference and aerodynamic stability. After comparing the results, the 10m cable length was identified as the most suitable tether length that maintained a safe distance between the \gls*{UAV} and the payload while ensuring minimal EMI interference as well as payload swing.

\section{Results}
After determining the optimal system characteristics, a comparative analysis needs to be run to verify if the system can acquire usable data. The result of the analysis compares two data sets: one from a fixed-wing survey using Cs-3 cesium vapour magnetometers on the Piper Navajo PA 31-325 CR platform conducted in 2013, and the other from a drone survey using a proprietary magnetometer on a \gls*{VTOL} fixed-wing \gls*{UAV} platform. The analysis revealed that both systems performed similarly in terms of signal strength and signal attenuation, with a positive correlation of about 98\% (signal ratio of 0.98). However, elevation changes had a lesser impact on the drone systems' signal strength, indicating more stability in varying terrain.

\textcolor{black}{For the comparative analysis, an appropriately mineral rich was selected, and a comprehensive analysis was conducted over a field size of 500m $\times$ 700m grids in the Minden area, located in the south of Ontario, Canada. The survey followed a flight pattern with a line spacing of 25m, distinctly narrower compared to traditional helicopter-borne aeromagnetic surveys which generally utilize a line spacing of 100m. The elevation for the flight was set at 85m above ground level. The fixed-wing \gls*{VTOL} \gls*{UAV} flights were performed in a single day with a ground reference magnetometer.}

\begin{figure*}[t]
    \centering
    \includegraphics[width=.95\textwidth]{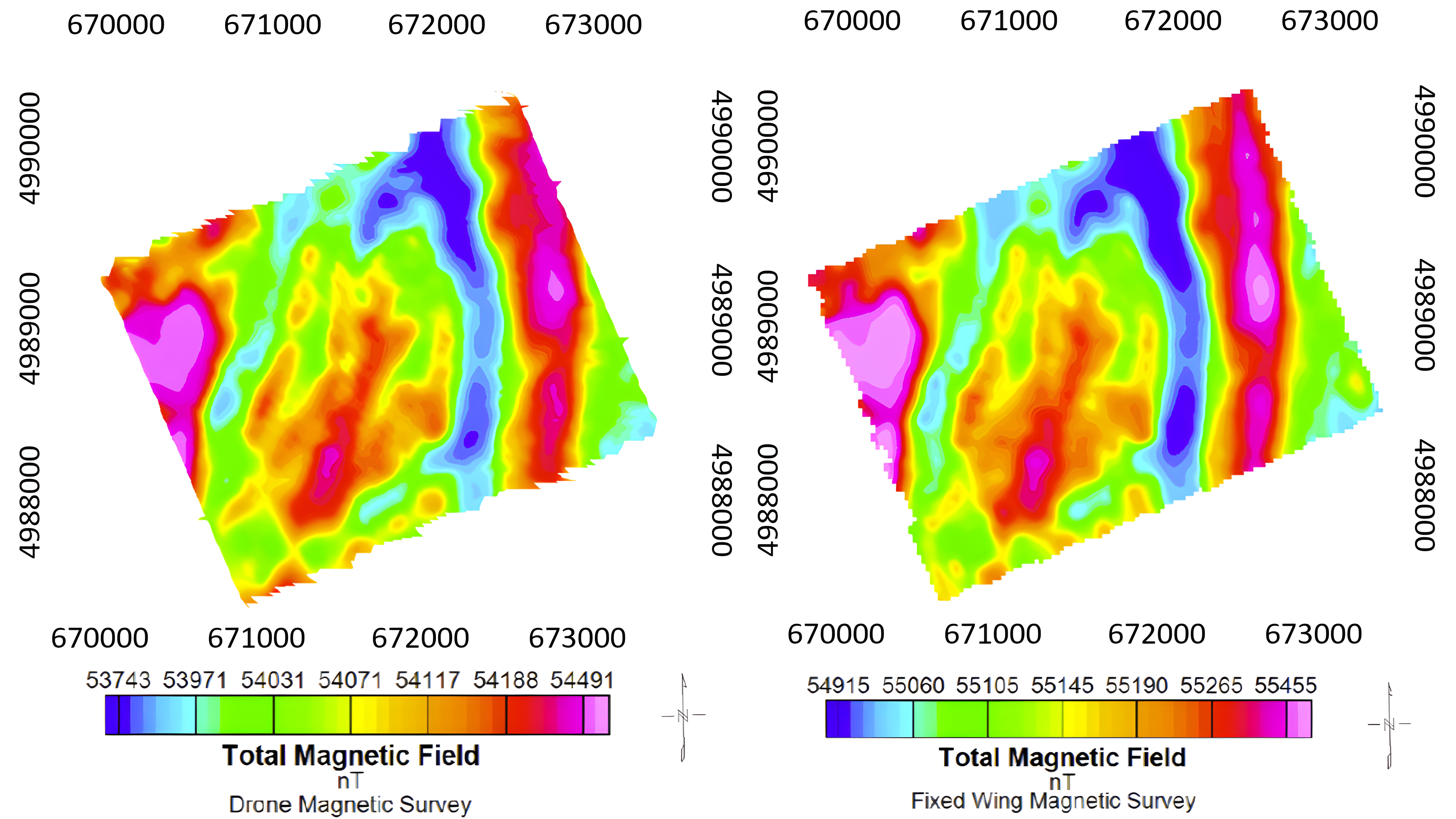}
    \caption[Fixed-wing manned vs UAV \gls*{TMI} map]{\textcolor{black}{Fixed-wing manned aircraft total magnetic intensity mapped (left) vs \gls*{VTOL} fixed-wing UAV total magnetic intensity (right).} \textcolor{black}{These two figures show that the fixed-wing \gls*{VTOL} survey over the same area provides equal or better resolution data compared to the manned survey. This is achieved while being safer and more economical in comparison. Numbers surrounding the plot indicate Eastings and Northings.\\}}
    \label{fig:Mag Results}
\end{figure*}
\vspace{5 mm}
\begin{figure*}[ht!]
    \centering
    \includegraphics[width=.95\textwidth]{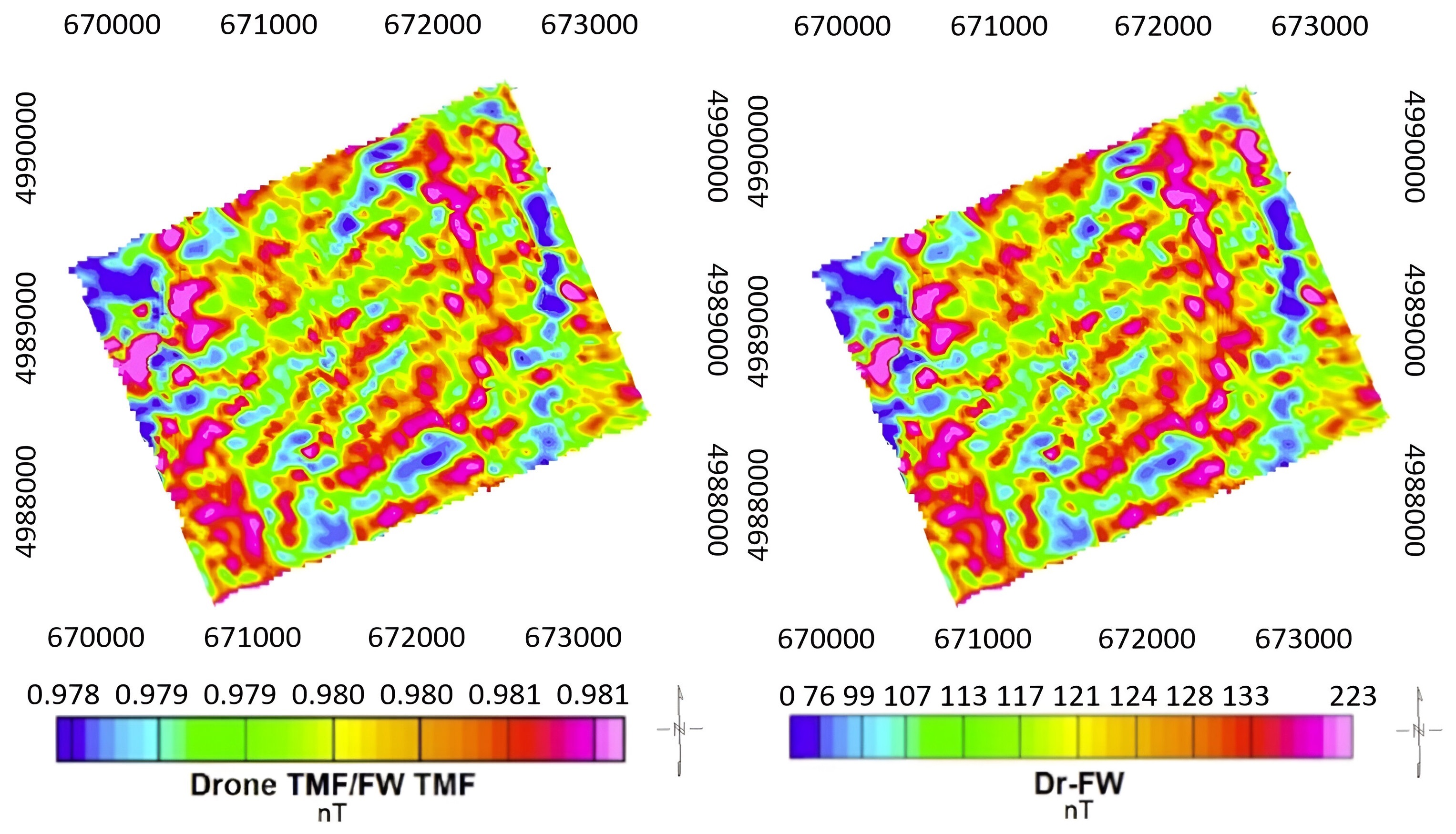}
    \caption[Ratio and subtraction based comparison]{\textcolor{black}{Ratio-based comparison of the signal strength between the drone system and manned fixed-wing system (top), as well as subtraction-based comparison (bottom).} \textcolor{black}{Drone Total magnetic frequency as a ratio compared to fixed-wing total magnetic frequency is represented as a nanotesla reading and the subtraction-based method of the drone data set minus the fixed-wing data set is outputted in nanoteslas as well. Numbers surrounding the plot indicate Eastings and Northings.}}
    \label{fig:Ratio Comparison}
\end{figure*}

\textcolor{black}{Fig.~\ref{fig:PiperVSdrone.jpg} shows the two systems used to perform the comparative analysis, both retrofitted with their own respective magnetometers for data acquisition. The Piper Navajo offers the advantage of covering large areas quickly and is particularly cost-effective for large-scale surveys with moderate data quality requirements. However, its high-altitude flights can lead to lower-resolution data, and its design poses some maneuverability challenges, especially in complex terrains. Meanwhile, \gls*{VTOL} fixed-wing \gls*{UAV}s deliver high-resolution data and are especially suited for smaller or intricate terrains, enhancing safety by eliminating the need for personnel in potentially hazardous areas. The decision between the two largely rests on the specific demands and scale of the survey in question.} After performing a set of flights to compare the data of each system, the noise analysis revealed that the drone system was on par with the fixed-wing platform in terms of high-frequency noise. To compensate for differences in line spacing between the two datasets, the drone data was re-sampled to approximate the fixed-wing data sampling intervals. Overall, while both systems showed comparable performance in signal strength and attenuation as seen in Fig.~\ref{fig:Mag Results}, the \gls*{UAV} system resulted in higher quality data due to its lower altitude.

\begin{figure*}[t!]
    \centering
    \includegraphics[width=.88\textwidth]{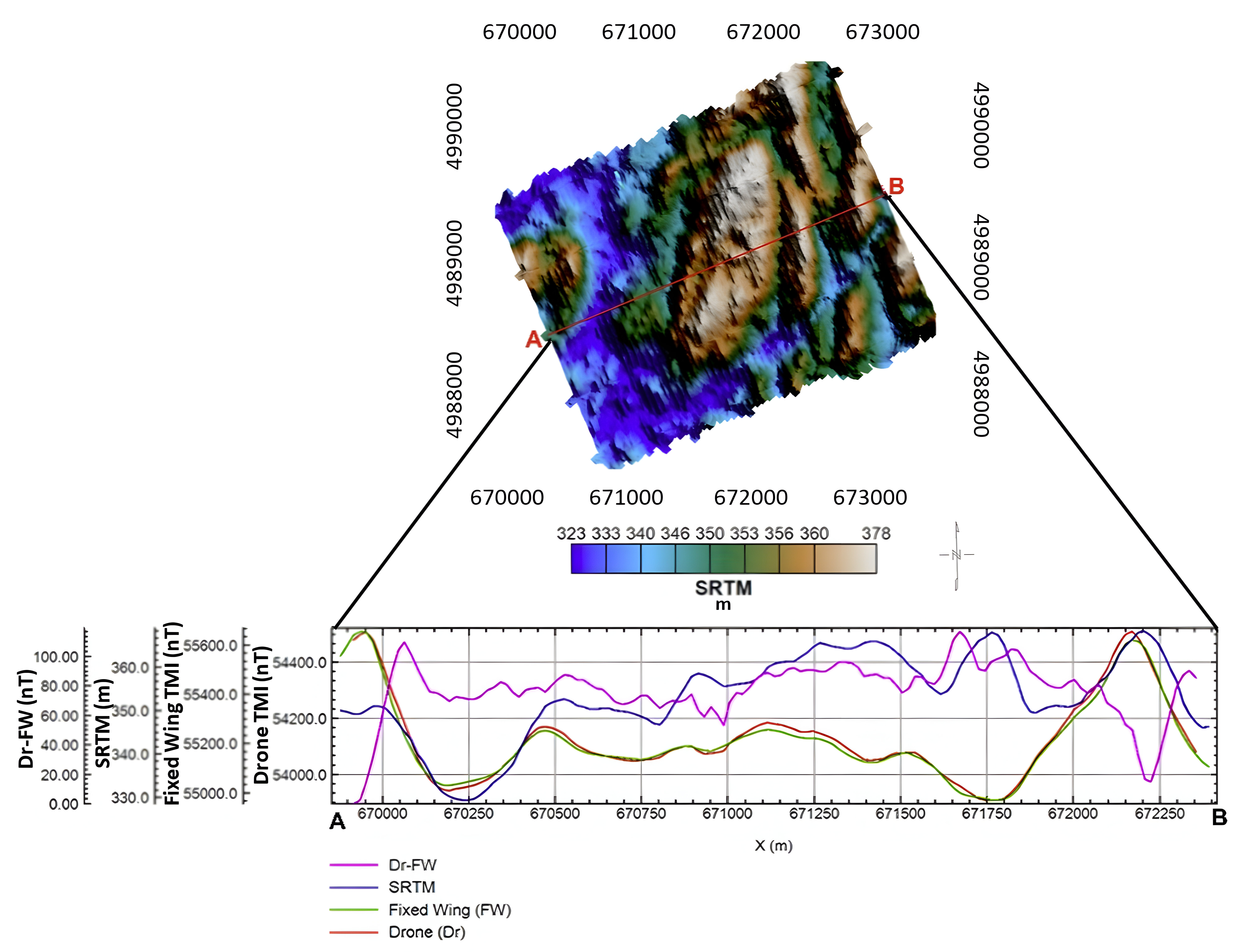}
    \caption[Signal strength to the SRTM graph]{\textcolor{black}{Graphing the observed relationship of variation in signal strength to the SRTM. The DR-\gls*{FW} plot indicates the difference in nT between the two readings, displaying higher performance in areas with lower elevation. The individual \gls*{TMI} of both sensors (in nT) are also plotted alongside the \gls*{SRTM} elevation (in m). Numbers surrounding the plot indicate Eastings and Northings.}}
    \label{fig:Srtm comparison}
\end{figure*}

Fig.~\ref{fig:Ratio Comparison} reveals the signal ratio between drone and fixed-wing data to be nearly 1, indicating both have similar capabilities when levelled at the same altitude.  \textcolor{black}{The total magnetic field (TMF) range is indicated at the bottom of the image showing the minimum and maximum nanotesla (nT) readings for that section of the flight. Statistical analysis was performed which in this case was focused on descriptive statistics to provide a summary of the main aspects of the data, such as mean, median, mode, and standard deviation. This analysis of the drone data led to a mean value of 1082 nT and a zero offset value of 116 nT, which was added to the drone total magnetic intensity vs fixed-wing total magnetic intensity difference to convert it into positive values for easier comparison. Normalization and calibration in data analysis, while beneficial, come with potential downsides such as the loss of absolute reference, potential misinterpretation, and the amplification of noise. In this study, meticulous steps were implemented to address these challenges and to minimize their effects, ensuring the results maintain both relevance and accuracy. This normalization process was required in order to properly perform the subtraction-based analysis but the unnormalized data can be found in Fig.~\ref{fig:Mag Results}. After applying these methods, results depicted in Fig.~\ref{fig:Ratio Comparison} confirmed that changes in signal largely stem from topographical variations, rather than differences in the acquisition systems' capabilities.} \textcolor{black}{\gls*{SRTM} was a collaborative project to generate high-resolution digital topographic maps of the Earth's surface that depict the height of the Earth's landforms above sea level.} The correlation between magnetic images and \gls*{SRTM} elevation data suggests that the observed changes in signal strength are also influenced by the area's geology. To confirm the idea that signals change is due to terrain, \textcolor{black}{a terrain profile} was extracted from both grids and contrasted with the observed difference between drone and fixed-wing signals. Fig.~\ref{fig:Srtm comparison} illustrates the location of the chosen profile and the variations in signal strength between the two systems in relation to SRTM-derived elevations.

Analyzing Fig.~\ref{fig:Srtm comparison}, both \gls*{UAV} and fixed-wing systems generally have equivalent signal measurement capability when levelled at the same altitude. However, distinctive variations occur in valleys and hills. The \gls*{UAV} system performs better in low-elevation areas, with only subtle differences detected from the fixed-wing dataset. In high-elevation areas, the difference nears zero, mainly because both systems are close to the magnetic source and the fixed-wing easily maintains terrain clearance. \textcolor{black}{The average variation in the calculated signal strength difference is around 60 nT to 80 nT, the typical range of an Overhauser magnetometer is from 20,000 to 100,000 nT for reference.} Over larger distances, both systems show minor variations with subtle topography changes with the \gls*{UAV} system being higher resolution at lower altitudes.

\textcolor{black}{Fig.~\ref{fig:Ratio Comparison} displays a signal ratio close to 1 between drone and fixed-wing data, suggesting equivalent capabilities at the same altitude. The drone data's statistical analysis highlighted a mean value of 1082 nT with a zero offset of 116 nT. This offset, when combined with the total magnetic intensity of both drone and fixed-wing data, transformed the values for straightforward comparison by means of descriptive statistics. Based on this, it is clear that drone-based systems for mineral data acquisition provide superior advantages in terms of cost while providing equal or greater quality.} The analysis results demonstrate that drones perform equally or, in certain scenarios, surpass the data quality obtained from traditional fixed-wing manned flights, approximately 10\% of the time for this particular dataset. Specifically, \gls*{UAV}s showcase superior sensor performance in areas of low elevation, which can be critical in mineral exploration. They also exhibit similar capabilities in signal strength and signal attenuation, providing accurate and reliable data sets. In addition to their high-quality data acquisition, drones offer substantial cost benefits. The elimination of crew requirements reduces both direct and indirect expenses, such as salaries, insurance, and training. The operational costs are further minimized due to lower fuel consumption and maintenance requirements. Finally, drones enhance safety, thereby mitigating risks associated with manned operations in challenging environments. These factors, combined with the data acquisition performance itself, position \gls*{VTOL} fixed-wing \gls*{UAV} technology as an attractive, cost-effective, and high-quality solution for mineral data acquisition, and a strong contender for future exploration projects.

Through a combination of tether optimization, \gls*{UAV} modifications, and payload adjustments, a system was developed which successfully employs the use of a suspended magnetometer payload for a \gls*{VTOL} fixed-wing aircraft. Experimental results demonstrated the effectiveness of this approach in reducing EMI interference and maintaining stable sensor operation during flight. The optimized tether length, coupled with the aerodynamic modifications to both the \gls*{UAV} and payload, resulted in high-quality mineral data acquisition with reduced noise levels. 

\section{Conclusions and Future Works}
\textcolor{black}{
The exploration of VTOL \gls*{UAV}s in the realm of mineral data acquisition has paved the way for efficient and accurate geological surveys. This research has provided insight into the challenges and solutions associated with integrating high-sensitivity sensors onto these platforms, specifically for VTOL configurations. The significance of this research lies in its novel approach to sensor integration on VTOL UAVs, addressing prevalent challenges like electromagnetic interference and payload dynamics which have been barriers in the past.} The main contribution of this research is the pioneering implementation of suspended magnetometry on a fixed-wing VTOL \gls*{UAV}. This innovative approach broadens the capabilities of \gls*{UAV}s in mineral exploration, presenting a shift from conventional fixed mounted methodologies in favour of a system with less noise. The uniqueness of this VTOL platform is further accentuated by a novel mounting scheme, optimizing sensor stability and performance. Such advancements not only underscore the technical ingenuity, but also its potential to set new standards in the field.

While \gls*{UAV}s have found application in various domains, their use in mineral exploration has predominantly been for basic aerial imagery and rudimentary data collection. This research, however, emphasizes the utility of VTOL \gls*{UAV}s in high-precision mineral data acquisition. The results of testing have shown the potential of these novel solutions in the field of \gls*{UAV}-based mineral data acquisition, advancing the domain of efficient mineral exploration. The findings herein underscore the importance and future potential of VTOL \gls*{UAV}s in revolutionizing mineral data collection.

The integration of advanced sensor systems onto VTOL \gls*{UAV} platforms, as illuminated in this research, introduces a new era in mineral data acquisition. This innovative stride prompts a natural progression towards enhancing sensor integration on future VTOL platforms. Building on the novel mounting scheme introduced in our study, the next phase of research could focus on the optimization of VTOL UAVs, with materials science potentially offering avenues for lighter, more resilient materials, ensuring enhanced sensor stability and longevity. 

Parallelly, the rich tapestry of data we accumulate from VTOL UAVs points towards the necessity of 3D data fusion techniques and analysis. The rapidly expanding fields of machine learning and artificial intelligence offer enticing prospects, potentially birthing models tailor-made for deciphering mineral data from VTOL \gls*{UAV}s. As we push the boundaries of VTOL \gls*{UAV} capabilities, safety and regulatory considerations must remain at the forefront, necessitating research into robust safety protocols and alignment with contemporary aviation regulations. To fully appreciate the nuances and efficacy of our advancements, empirical studies juxtaposing with new VTOL \gls*{UAV} mounted fluxgate magnetometers could illuminate relative efficiencies and accuracy's, providing a more detailed view of the strides made in mineral exploration.

\bibliographystyle{IEEEtran}

\bibliography{interactcadsample}  

\begin{thebibliography}{10}
\providecommand{\url}[1]{#1}
\csname url@samestyle\endcsname
\providecommand{\newblock}{\relax}
\providecommand{\bibinfo}[2]{#2}
\providecommand{\BIBentrySTDinterwordspacing}{\spaceskip=0pt\relax}
\providecommand{\BIBentryALTinterwordstretchfactor}{4}
\providecommand{\BIBentryALTinterwordspacing}{\spaceskip=\fontdimen2\font plus
\BIBentryALTinterwordstretchfactor\fontdimen3\font minus \fontdimen4\font\relax}
\providecommand{\BIBforeignlanguage}[2]{{%
\expandafter\ifx\csname l@#1\endcsname\relax
\typeout{** WARNING: IEEEtran.bst: No hyphenation pattern has been}%
\typeout{** loaded for the language `#1'. Using the pattern for}%
\typeout{** the default language instead.}%
\else
\language=\csname l@#1\endcsname
\fi
#2}}
\providecommand{\BIBdecl}{\relax}
\BIBdecl

\bibitem{mohsan2023editorial}
S.~A.~H. Mohsan, M.~A. Khan, and Y.~Y. Ghadi, ``Editorial on the advances, innovations and applications of {UAV} technology for remote sensing,'' \emph{Remote Sensing}, vol.~15, p. 5087, 2023.

\bibitem{vangu2022use}
G.~M. Vangu, ``The use of drones in mining operations,'' \emph{Mining Revue}, vol.~28, no.~3, pp. 73--82, 2022.

\bibitem{shahsavani2021aeromagnetic}
H.~Shahsavani, ``An aeromagnetic survey carried out using a rotary-wing {UAV} equipped with a low-cost magneto-inductive sensor,'' \emph{International Journal of Remote Sensing}, vol.~42, no.~23, pp. 8805--8818, 2021.

\bibitem{leech2021acquisition}
C.~Leech, S.~Burns, and K.~Hurley, ``Acquisition challenges for high quality data using a {UAV} deployed magnetometer,'' in \emph{Sixth International Conference on Engineering Geophysics}, Virtual, 2021, oct. 25--28, 2021.

\bibitem{tziavou2018unmanned}
O.~Tziavou, S.~Pytharouli, and J.~Souter, ``{Unmanned Aerial Vehicle} ({UAV}) based mapping in engineering geological surveys: Considerations for optimum results,'' \emph{Engineering Geology}, vol. 232, pp. 12--21, 2018.

\bibitem{lu2023development}
N.~Lu, Y.~Xi, H.~Zheng, W.~Gao, Y.~Li, Y.~Liu, Z.~Cui, G.~Liao, and J.~Liu, ``Development of a hybrid fixed-wing {UAV} aeromagnetic survey system and an application study in chating deposit,'' \emph{Minerals}, vol.~13, p. 1094, 2023.

\bibitem{giordan2020use}
D.~Giordan, M.~S. Adams, I.~Aicardi, M.~Alicandro, P.~Allasia, M.~Baldo, P.~De~Berardinis, D.~Dominici, D.~Godone, P.~Hobbs, V.~Lechner, T.~Niedzielski, M.~Piras, M.~Rotilio, R.~Salvini, V.~Segor, B.~Sotier, and F.~Troilo, ``The use of {Unmanned Aerial Vehicle}s ({UAV}s) for engineering geology applications,'' \emph{Bulletin of Engineering Geology and the Environment}, vol.~79, no.~7, pp. 3437--3481, 2020.

\bibitem{vitale2019new}
G.~Vitale, S.~Scudero, A.~D'Alessandro, A.~Pisciotta, R.~Martorana, and P.~Capizzi, ``New ultraportable data logger to perform magnetic surveys,'' in \emph{2019 International Symposium on Advanced Electrical and Communication Technologies (ISAECT)}, 2019.

\bibitem{mat2023magnetic}
M.~Mat, ``Magnetic surveys: Principles, applications \& geology science,'' \url{https://geologyscience.com/geology-branches/geophysics/magnetic-surveys/}, 2023, accessed on: Dec. 3, 2023.

\bibitem{persova2021resolution}
M.~G. Persova, Y.~G. Soloveichik, D.~V. Vagin, D.~S. Kiselev, A.~P. Sivenkova, and E.~I. Simon, ``Resolution analysis of airborne electromagnetic survey using helicopter platform and {UAV},'' in \emph{2021 XV International Scientific-Technical Conference on Actual Problems Of Electronic Instrument Engineering (APEIE)}, Novosibirsk, Russian Federation, 2021, pp. 591--594.

\bibitem{jiang2020integration}
D.~Jiang, Z.~Zeng, S.~Zhou, Y.~Guan, and T.~Lin, ``Integration of an aeromagnetic measurement system based on an {Unmanned Aerial Vehicle} platform and its application in the exploration of the ma’anshan magnetite deposit,'' \emph{IEEE Access}, vol.~8, pp. 189\,576--189\,586, 2020.

\bibitem{greengard2019when}
S.~Greengard, ``When drones fly,'' \emph{Communications of the ACM}, vol.~62, no.~11, pp. 16--18, 2019.

\bibitem{nex2014uav}
F.~Nex and F.~Remondino, ``{{UAV}} for 3d mapping applications: A review,'' \emph{Applied Geomatics}, vol.~6, no.~1, pp. 1--15, 2014.

\bibitem{akshat2022review}
M.~Akshat, S.~Jayachandran, S.~Kenche, A.~Katoch, A.~Suresh, E.~Gundabattini, S.~K. Selvaraj, and A.~A. Legesse, ``A review on vertical take-off and landing {(VTOL)} tilt-rotor and tilt wing {Unmanned Aerial Vehicle}s {({UAV}s)},'' \emph{Journal of Engineering}, p. 1803638, 2022.

\bibitem{padua2017uas}
L.~Pádua, J.~Vanko, J.~Hruška, T.~Adão, J.~J. Sousa, E.~Peres, and R.~Morais, ``{UAS}, sensors, and data processing in agroforestry: A review towards practical applications,'' \emph{International Journal of Remote Sensing}, vol.~38, no. 8--10, pp. 2349--2391, 2017.

\bibitem{barton2021extending}
I.~F. Barton, M.~J. Gabriel, J.~Lyons-Baral \emph{et~al.}, ``Extending geometallurgy to the mine scale with hyperspectral imaging: a pilot study using drone- and ground-based scanning,'' \emph{Mining, Metallurgy \& Exploration}, vol.~38, pp. 799--818, 2021.

\bibitem{niethammer2012uav}
U.~Niethammer, M.~R. James, S.~Rothmund, J.~Travelletti, and M.~Joswig, ``{{UAV}}-based remote sensing of the super-sauze landslide: Evaluation and results,'' \emph{Engineering Geology}, vol. 128, pp. 2--11, 2012.

\bibitem{salvini2015geological}
R.~Salvini, S.~Riccucci, D.~Gullì, R.~Giovannini, C.~Vanneschi, and M.~Francioni, ``Geological application of {{UAV}} photogrammetry and terrestrial laser scanning in marble quarrying (apuan alps, italy),'' in \emph{Engineering Geology for Society and Territory - Volume 5}, G.~Lollino, A.~Manconi, F.~Guzzetti, M.~Culshaw, P.~Bobrowsky, and F.~Luino, Eds.\hskip 1em plus 0.5em minus 0.4em\relax Springer, 2015, pp. 1883--1887.

\bibitem{jackisch2019drone}
R.~Jackisch, Y.~Madriz, R.~Zimmermann, M.~Pirttijärvi, A.~Saartenoja, B.~H. Heincke, H.~Salmirinne, J.-P. Kujasalo, L.~Andreani, and R.~Gloaguen, ``Drone-borne hyperspectral and magnetic data integration: Otanmäki fe-ti-v deposit in finland,'' \emph{Remote Sensing}, vol.~11, no.~18, p. 2084, 2019.

\bibitem{shendryk2020fine}
Y.~Shendryk, J.~Sofonia, R.~Garrard, Y.~Rist, D.~Skocaj, and P.~Thorburn, ``Fine-scale prediction of biomass and leaf nitrogen content in sugarcane using {{UAV}} lidar and multispectral imaging,'' \emph{International Journal of Applied Earth Observation and Geoinformation}, vol.~92, p. 102177, 2020.

\bibitem{dundar2020design}
O.~Dundar, M.~Bilici, and T.~Unler, ``Design and performance analyses of a fixed wing battery {VTOL} {{UAV}},'' \emph{Engineering Science and Technology, an International Journal}, vol.~23, no.~5, pp. 1182--1193, 2020.

\bibitem{shahmoradi2020comprehensive}
J.~Shahmoradi, E.~Talebi, P.~Roghanchi, and M.~Hassanalian, ``A comprehensive review of applications of drone technology in the mining industry,'' \emph{Drones}, vol.~4, no.~3, p.~34, 2020.

\bibitem{goetz2009three}
A.~F. Goetz, ``Three decades of hyperspectral remote sensing of the earth: A personal view,'' \emph{Remote Sensing of Environment}, vol. 113, pp. S5--S16, 2009.

\bibitem{vosselman2013recognising}
G.~Vosselman, B.~Gorte, G.~Sithole, and T.~Rabbani, ``Recognising structure in laser scanner point clouds,'' in \emph{International Archives of the Photogrammetry, Remote Sensing and Spatial Information Sciences}, vol.~46, 2013, pp. 33--38.

\bibitem{thiele2021multi}
S.~T. Thiele, S.~Lorenz, M.~Kirsch, I.~C.~C. Acosta, L.~Tusa, E.~Herrmann, R.~Möckel, and R.~Gloaguen, ``Multi-scale, multi-sensor data integration for automated 3-d geological mapping,'' \emph{Ore Geology Reviews}, vol. 136, p. 104252, 2021.

\bibitem{molnar2016unmanned}
A.~Molnar and C.~Parsons, ``{Unmanned Aerial Vehicle}s ({{UAV}}s) and law enforcement in australia and canada: Governance through ‘privacy’ in an era of counter-law?'' in \emph{National Security, Surveillance and Terror}, R.~Lippert, K.~Walby, I.~Warren, and D.~Palmer, Eds.\hskip 1em plus 0.5em minus 0.4em\relax Palgrave Macmillan, 2016, pp. 183--200.

\bibitem{egbert2020modelling}
G.~D. Egbert, P.~Alken, A.~Maute, and H.~Zhang, ``Modelling diurnal variation magnetic fields due to ionospheric currents,'' \emph{Geophysical Journal International}, vol. 225, no.~2, pp. 1086--1109, 2020.

\bibitem{inverse-square-law-2014}
\BIBentryALTinterwordspacing
(2014) Inverse square law, general. \url{http://hyperphysics.phy-astr.gsu.edu/hbase/Forces/isq.html}. Accessed: 3-Jan-2014. [Online]. Available: \url{http://hyperphysics.phy-astr.gsu.edu/hbase/Forces/isq.html}
\BIBentrySTDinterwordspacing

\bibitem{Halliday2021}
D.~Halliday, R.~Resnick, and J.~Walker, \emph{Fundamentals of Physics}, 12th~ed.\hskip 1em plus 0.5em minus 0.4em\relax John Wiley \& Sons, Inc., 2021.

\bibitem{tmotor2023at7224}
{T-Motor}. (n.d.) {AT7224 40CC. AT7224 40cc\_at series\_motors\_fixed wing\_t-motor store-official store for T-Motor Drone Motor,ESC,Propeller}. \url{https://store.tmotor.com/goods-1094-AT7224+40CC.html}.

\bibitem{willems2009}
M.~Willems. (2009, July 12) {C-GJBG: Piper PA-31-310 Navajo C: Goldak airborne surveys: Matt Willems}. \url{https://www.jetphotos.com/photo/6611557}.

\end{thebibliography}

\newpage
\onecolumn
\appendix
\begingroup
    \let\clearpage\relax
    \vspace*{2\baselineskip}
\section{(Derivations)}
\label{appendixA}
\onecolumn
\subsection{Calculations for Mineral Data Acquisition Methods}
\vspace*{1\baselineskip}
\vspace{4 mm}
\noindent\textbf{Cost per Line Km for Generic UAV:}
Using the Generic UAV as an example, calculating the `Cost per Line Km' as follows. First, the number of flights possible in an 8-hour workday then the operating costs per flight:
\begin{equation}
    \text{Number of Flights per Day} = \frac{\text{Total Work Hours per Day}}{\text{Flight Duration (hrs)}} = \frac{8}{0.5} = 16 \text{ flights}
\end{equation}
\begin{equation}
    \text{Operating Costs per Flight} = \frac{\text{Average Operating Costs per Day}}{\text{Number of Flights per Day}} = \frac{500}{16} = \$31.25
\end{equation}

\noindent Finally, calculate the distance covered in one flight and the final cost per Line Km:
\begin{equation}
    \begin{split}
        \text{Distance Covered per Flight (km)} &= \text{Flight Duration (hrs)} \times \text{Flight Speed (km/hr)} \\
        &\phantom{{}=} 0.5 \times 45 = 22.5 \text{ km}
    \end{split}
\end{equation}
\begin{equation}
    \text{Cost per Line Km} = \frac{\text{Operating Costs per Flight}}{\text{Distance Covered per Flight (km)}} = \frac{31.25}{22.5} \approx \$1.38 \text{ per km}
\end{equation}

\vspace{4 mm}
\noindent This calculation shows that, under these assumptions, the Cost per Line Km for the Generic UAV is approximately \$1.38 per km. Adjusting the flight speed or other parameters will change this value accordingly.

\vspace*{2\baselineskip}
\vspace{4 mm}
\noindent\textbf{Coverage Area per Flight (km\(^2\)):}
Dependent on flight duration, speed, and effective coverage width. For example, a generic UAV covering a 0.025 km wide area in a single pass, flying for 0.5 hours at 40 km/h, would have:
\begin{equation}
    \text{Coverage Area} = \text{Width} \times \text{Flight Duration} \times \text{Speed} = 0.025 \times 0.5 \times 40 = 0.5 \text{ km}^2
\end{equation}

\vspace*{2\baselineskip}
\vspace{4 mm}
\noindent\textbf{Average Operating Costs (\$ per day):}
Includes fuel costs, maintenance, and crew expenses. For example, if the fuel cost is \$50, maintenance is \$100, and crew costs are \$350, the total would be \$500 per day. Of course, additional expenses are to be expected and costs may change slightly but the simplified values are represented.

\vspace*{2\baselineskip}
\vspace{4 mm}
\noindent\textbf{Flight Time per km (seconds/km):}
Calculated based on the speed of the UAV. For a UAV traveling at 40 km/h:
\begin{equation}
    \text{Flight Time per km} = \frac{1 \text{ hour}}{\text{Speed (km/h)}} \times 3600 \text{ seconds} = \frac{1}{40} \times 3600 = 90 \text{ seconds/km}
\end{equation}

\endgroup

\end{document}